# APPLICATION OF VISION-LANGUAGE MODELS FOR ASSESSING OSTEOARTHRITIS DISEASE SEVERITY

*Banafshe Felfeliyan, Yuyue Zhou, Shrimanti Ghosh, Shaobo Liu,*
Abhilash Hareendranathan, Jacob L. Jaremko

Department of Radiology & Diagnostic Imaging, University of Alberta, Edmonton, AB, Canada

## ABSTRACT

Osteoarthritis (OA) poses a global health challenge, demanding precise diagnostic methods. Current radiographic assessments are time-consuming and prone to variability, prompting the need for automated solutions. The existing deep learning models for OA assessment are unimodal single-task systems and they don't incorporate relevant text information such as patient demographics, disease history, or physician reports. This study investigates employing Vision-Language Processing (VLP) models to predict OA severity using X-ray images and corresponding reports. Our method leverages X-ray images of the knee and diverse report templates generated from tabular OA scoring values to train a CLIP (Contrastive Language-Image Pre-Training) style VLP model. Furthermore, we incorporate additional contrasting captions to enforce the model to discriminate between positive and negative reports. Results demonstrate the efficacy of these models in learning text-image representations and their contextual relationships, showcase potential advancement in OA assessment, and establish a foundation for specialized vision-language models in medical contexts. ***Index Terms—***

Vision-Language, Multi-modal Representation Learning, Osteoarthritis, Radiology Report

## 1. INTRODUCTION

Osteoarthritis (OA), a progressive degenerative disease, stands as a leading cause of pain and disability among adults, causing a substantial burden on public health [1]. Moreover, OA holds the distinction of being the most prevalent joint disease worldwide, affecting over 500 million people [2] and the number of people affected worldwide has increased by 48% from 1990 to 2019 [2]. Imaging plays an important role in the diagnosis and feature characterization of OA [3]. Currently, the common clinical approach to diagnose and monitor the long-term progression is measuring the joint space narrowing (JSN) and assessing OA features using the Kellgren-Lawrence (KL) scoring in the planar radiograph (X-ray) [4].

However, KL grades are subject to intra/inter-reader variability and take a considerable amount of time which can have significant implications for patients and their diagnoses. Therefore, the need for an automatic method for evaluating X-rays of individual OA patients has become more evident over the last few years [5]. Such a method would facilitate more precise tracking of their disease progression.

Over the past decade, different studies have deployed and developed numerous deep-learning (DL) methods to assess the severity of OA in different imaging modalities [6–8]. Different approaches have been proposed to predict KL grade from the X-ray images using deep-deep learning and generated relatively accurate results [5].

Despite the progress of DL methods, most of the models that are developed for OA assessment or other biomedical domains are unimodal single-task systems and they cannot incorporate relevant information such as patient demographics and disease history [9], which can bound their performance in real-world applications.

To address this problem, different studies have begun to incorporate Vision-Language Processing (VLP) models, which integrate both visual and textual information, using medical images and corresponding text reports data [10–12]. Vision-language models have explored their effectiveness on a variety of downstream tasks [9, 13] and have demonstrated that cross-modal supervision can offer a more robust signal for training both images [14] and text [15] models, demonstrating considerable potential in various medical applications. The majority of methods primarily concentrate on vision-language pre-training specifically for chest X-rays. The utilization of these models holds promise in health, particularly in the realm of assessing osteoarthritis (OA) disease severity, presenting an area ripe for their effective application. To the best of our knowledge, there is no work for a multi-modal DL model in the OA domain.

Motivated by these observations, in this paper, we propose to deploy a CLIP-style VLP model to predict the severity of OA in X-ray images. The proposed method is developed based on tabular medical scores data by increasing the data efficiency during vision-language modeling by deploying multiple caption styles. Additionally, the proposed method introduces an OA X-ray domain-specific vision-language pre-

———————————
Corresponding author: B.F., email: banfel@ualberta.ca

trained model that can be used in other different supervised and unsupervised downstream tasks. Our contributions can be summarized as follows:

- We investigate CLIP's potential with OA images and assess its zero-shot transfer capabilities.
- We generate diverse radiology report templates derived from tabular scoring values.
- Our proposal involves employing additional contrasting captions to prompt the network to diminish the similarity between positive and negative reports by reducing the cosine similarity within their embedding space.

## 2. RELATED WORKS (VISION-TEXT REPRESENTATION LEARNING IN THE MEDICAL DOMAIN)

The availability of large-scale paired image-text datasets presented a significant opportunity for the fast development of VLP models to develop rapidly for general-purpose applications with different supervised and self-supervised objectives [16]. CLIP [17] is known as one of the groundbreaking large-scale vision-language pretraining (VLP) models, which utilized contrast learning to learn image representations on a dataset of 400 million pairs (image, text) from scratch. It can be used for image-text similarity and for zero-shot image classification. CLIP uses a ViT-like transformer to get visual features and a causal language model to get text features.

In the medical field, some research has incorporated principles from general-domain Vision-Language Pre-trained (VLP) models by harnessing automated label extraction techniques from clinical reports, including the CLIP-style cross-modal contrastive objective [18].

Wang et al. introduced MedCLIP [12], diverging from the CLIP model by incorporating a semantic matching loss instead of InfoNCE. This alteration aims to treat text as a soft target, reducing false negatives. MedCLIP's training involves 570,000 image-text pairs and requires supplementary annotated data to define its objective function.

CLIPath proposed by Lai et al, [19] was proposed to enhance the classification accuracy of pathology images using limited labeled data. This approach integrates a Residual Feature Connection (RFC) to fine-tune CLIP, effectively optimizing with a minimal number of trainable parameters. Stember et al. [20], trained an SBERT (sentence BERT) using a limited number of radiology reports. They subsequently employed the reports generated by the trained SBERT model and 40 images to train an RL-based classifier. They demonstrated achieving a higher accuracy using with using language and vision representation.

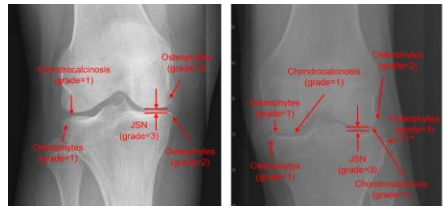

Fig. 1. Examples of knee OA with different KL, JSN, and osteophyte grades. JSN is assessed separately for lateral and medial compartments. Osteophytes are assessed separately

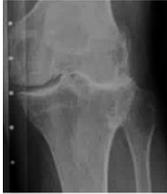

Fig. 2. X-ray image and example of 3 corresponding captions 1) based on abnormality, 2) based on location, and 3) overall insight

## 3. MATERIAL AND METHODS

### 3.1. Data

Data were retrieved from the publicly available Osteoarthritis Initiative (OAI) dataset. OAI dataset is a multicenter, longitudinal, prospective observational study of 4796 individuals with or at risk of developing knee OA. OAI contains bilateral fixed flexion plain X-ray images of the knee, as well as demographic and clinical information with, and KL and OARSI scores. The KL grading incorporates multiple features to establish an OA score of 0-4, and is the current accepted standard scoring method for radiographic knee OA diagnosis [21]. Furthermore, this study used the Osteoarthritis Research Society International (OARSI) atlas criteria scoring which grades various aspects of osteoarthritis [22] (Fig. 1).

This study utilized 5204 bilateral X-ray images from 2604 subjects, each paired with their respective KL grade and OARSI scores, resulting in the generation of 10408 single-limb images.

#### 3.1.1. Caption Preparation

The patient's data and OA scores are stored in a numerical tabular format. Given the debates on the ability of language models to understand measurements [23] in this study, we only used qualitative scoring results and mapped them into descriptive words (example: 0=no, 1=early, 2=mild, 3=moderate, and 4=severe). This study used patient age and sex, knee alignment, KL grad, OARSI grades (for attrition, scle-rosis, and osteophytes) as well as the presence of cysts and chondrocalcinosis.

For each image, distinct sets of caption templates were designed in order to obtain multiple captions for each. The captions are produced based on three templates, 1) based on abnormality, 2) based on location, and 3) overall insight. During the training process, for each image, the corresponding caption was randomly selected from the bag of captions. Then another view of the selected caption was augmented using sentence shuffling [24]. An example of different captions is presented in Fig. 2.

For designing the negative captions for each score, we randomly selected scores that were at least 2 levels different than the positive scores. The reason for selecting the 2-level threshold instead of the 1-level threshold is to reduce the readers' uncertaintydiscrepancy effect.

### 3.2. Vision-Language Processing

Similar to CLIP, the proposed model is constructed from a Text encoder, an Image encoder, and a projection head. During training for each image, one caption is randomly selected from the caption-bag. Corresponding text and images are forwarded to text and image encoders to obtain image embedding (= $I$) and text embedding's (= $T$). Next, it computes cosine similarities between projected Image and Text embedding's is calculated and is used to calculate an InfoLoss adapted for contrastive (text, image) representation [16]. In this paper, we proposed the incorporation of an additional loss function to reduce the similarity between each caption and its corresponding negative version, by minimizing the cosine similarity between the un-projected embeddings of these paired captions.

In this study, a pre-trained CXR-BERT-general from the [15] was used as the backbone text encoder, which was pre-trained on PubMed abstracts and clinical notes from two other publicly available datasets (MIMIC-IIICXR). Furthermore, the ResNet50 [25] was used to serve as the image encoder.

### 3.3. Implementation Details

In this study a pre-trained CXR-BERT-general from the [15] was used as the backbone text encoder which was pre-trained on PubMed abstracts and clinical notes from two other publicly available datasets.

From the 10400 image instances, 8,430 were used for training, 936 for validation, and 1,040 for testing. In each epoch, we randomly excluded training images with similar scores (captions) to avoid competitive contrastive learning among similar examples, which yielded 7500 images for training.

The proposed method was implemented in Pytorch and trained on an NVIDIA V100 GPU using Compute Canada resources. The dimensions of input images were configured to 224 × 224. We limited the maximum epochs to 20 with batch size 32 and used the Adam optimizer with a weight decay parameter of 1e-3. The learning rate for the image encoder, text encoder, and projection head respectively was 1e-4, 1e-6, and 1e-3.

## 4. EXPERIMENTS AND RESULTS

### 4.1. Model Interpretation

Gradient-weighted Class Activation Mapps (GRAD-CAM) [26] were generated (Fig. 3) to interpret the model's capacity in associating text with image features. Maps highlight crucial regions influenced by text input, showcasing CLIP's attempt to identify diverse OA features and employ comprehensive image details. In particular, intercondylar tubercles and joint margins are prominently featured on activation maps, indicating the model's ability to link textual prompts to visual representations.

### 4.2. Image classification

The accuracy of the proposed pertained image encoders for zero-shot classification on KL grade scoring was 55% which is 10% higher than the original CLIP model.

### 4.3. Image-Text Retrieval

In this experiment, textual information was retrieved for each image query in the test set, few examples are presented in Fig. 4. Quality of the retrieved captions was evaluated by BLEU [27] which is an established metric commonly used in image captioning. The BLEU4 value of the test dataset in our proposed method was 0.7 and for the original CLIP model 0.67.

## 5. DISCUSSION AND CONCLUSION

This paper explored the application of VLP models on OA images, transforming tabular data into various styles of text captions. Activation maps highlighted the areas related to OA features. This is a showcase of the capability of these models to establish connections between contextual (scoring) representations in OA X-ray images. Additionally, the introduced additional loss function enhanced zero-shot classification by compelling the model to create separation between positive and negative captions in embedding space. The research highlighted VLP's potential to align visual and textual data, promising improved diagnostic accuracy and personalized treatment approaches.

## 6. ACKNOWLEDGMENTS

Academic time for JJ is made available by Medical Imaging Consultants (MIC), Edmonton, Canada. JJ is supported by a Canada CIFAR AI Chair. BF is supported by an Alberta Innovates scholarship.

**Fig. 3**. Examples of activation maps for different given prompts. Results show model emphasis on the medial and lateral joint margins, as well as the intercondylar tubercles, indicating that the model was able to relate text and image representations.

**Fig. 4**. Example of Retrieved captions for the quarry images